\documentclass[%
reprint,
superscriptaddress,
floatfix,
amsmath,
amssymb,
aps,
notitlepage
]{revtex4-1}

\usepackage{placeins}

\usepackage[hidelinks=true]{hyperref}
\usepackage{xcolor}
\usepackage{threeparttable}
\usepackage{comment}

\definecolor{greenish}{RGB}{108, 200, 105}
\definecolor{reddish}{RGB}{174,12,48}
\definecolor{blueish}{rgb}{0.12, 0.56, 1.0}
\definecolor{magenta}{RGB}{242, 80, 93}
\hypersetup{
  colorlinks   = true,
  urlcolor     = greenish,
  linkcolor    = blueish,
  citecolor    = magenta
}

\usepackage{amsmath}
\usepackage{setspace}
\usepackage{caption}
\usepackage{float}
\usepackage{siunitx}
\usepackage[utf8]{inputenc}
\usepackage{subcaption}
\usepackage{graphicx}
\usepackage{dcolumn}
\usepackage{bm}

\makeatletter
\newcommand*{\rom}[1]{\exp\!andafter\@slowromancap\romannumeral #1@}
\makeatother

\def\d{\text{d}}
\def\A{\mathcal{A}}

\def\E{\mathcal{E}}
\def\F{\mathcal{F}}
\def\G{\mathcal{G}}
\def\S{\mathcal{S}}
\def\W{\mathcal{W}}
\def\K{\mathcal{K}}
\def\r{\mathbf{r}}
\def\f{\mathbf{f}}
\def\m{\mathbf{m}}
\def\p{\mathbf{p}}
\def\q{\mathbf{q}}
\def\rho{\varrho}
\def\beta{k_a}
\def\lambda{k_s}

\def\that{\hat{\mathbf{t}}}
\def\nhat{\hat{\mathbf{n}}}

\def\ez{\hat{\mathbf{e}}_z}
\def\alpha{\mathcal{D}}
\def\eff{\mathrm{eff}}

\begin{document}
\title{Morphologies of a sagging \textit{elastica} with intrinsic sensing and actuation}
\author{Vishnu  Deo  Mishra}
\email{am24r007@smail.iitm.ac.in}
\affiliation{Department of Applied Mechanics \& Biomedical Engineering, IIT Madras, Chennai, TN 600036.}
\author{S  Ganga  Prasath}
\affiliation{Department of Applied Mechanics \& Biomedical Engineering, IIT Madras, Chennai, TN 600036.}

\begin{abstract}
The morphology of a slender soft-robot can be modified by sensing its shape via sensors and exerting moments via actuators embedded along its body. The actuating moments required to morph these soft-robots to a desired shape are often difficult to compute due to the geometric non-linearity associated with the structure, the errors in modeling the experimental system, and the limitations in sensing and feedback/actuation capabilities. In this article, we explore the effect of a simple feedback strategy (actuation being proportional to the sensed curvature) on the shape of a soft-robot, modeled as an \textit{elastica}. The finite number of sensors and actuators, often seen in experiments, is captured in the model via filters of specified widths. Using proportional feedback, we study the simple task of straightening the device by compensating for the sagging introduced by its self-weight. The device undergoes a hierarchy of morphological instabilities defined in the phase-space given by the gravito-bending number, non-dimensional sensing/feedback gain, and the scaled width of the filter. For complex shape-morphing tasks, given a perfect model of the device with limited sensing and actuating capabilities, we find that a trade-off arises (set by the sensor spacing \& actuator size) between capturing the long and short wavelength features. We show that the error in shape-morphing is minimal for a fixed filter width when we choose an appropriate actuating gain (whose magnitude goes as a square of the filter width). Our model provides a quantitative lens to study and design slender soft devices with limited sensing and actuating capabilities for complex maneuvering applications.
\end{abstract}

\maketitle
\section{\label{Introduction}Introduction}
Soft-robotic systems have shown promise in applications where safety and versatility are pivotal~\cite{Polygerinos_2015, Zhou_2015, Yang_2015, Chang_2021, Mahvash_2011, Bosi_2015, Cianchetti_biomedical_2018, Trivedi_2008, Webster_design_2010, schmitt_soft_2018, Santina_2023, Wang_cable-driven_2017, Runciman_soft_2019, Gifari_review_2019}, attributes arising primarily out of their compliance. They have found utility in personal robots that operate safely around humans~\cite{Polygerinos_2015}, service robots designed for high-dexterity tasks in confined spaces~\cite{Zhou_2015, Yang_2015, Chang_2021}, and medical robots for minimally invasive procedures such as endoscopic surgeries~\cite{Mahvash_2011, Bosi_2015, Cianchetti_biomedical_2018, Wang_cable-driven_2017, Runciman_soft_2019, Gifari_review_2019}. These soft systems often use living organisms as their inspiration, both for their morphologies~\cite{shepherd2011multigait, wehner2016integrated, rafsanjani2018kirigami} and actuation strategies~\cite{kaczmarski2024minimal, leanza2024elephant, gazzola2015gait, charles2025topological, hoffmann2024postural}, as the living system provides a confined functional morphological space as well as control protocols with few tuning parameters to perform a specific task.

Despite significant progress in fabrication and theoretical understanding of these devices in the recent years \cite{ranzani_increasing_2018,Thuruthel_control_2018,naughton_elastica_2021,mccandless_soft_2022,chang_energy-shaping_2023,leanza_elephant_2024,kaczmarski_minimal_2024,lee_fabrication_2024,wang_sensing_2024} translating these soft robots to real-world applications remains a challenge due to three dominant reasons: $(i, ii)$ retrieving spatio-temporal information about the state/shape of the soft device and providing the device with programmable shape change capability is difficult. This is due to the fact that the device can deform continuously thus demanding a large number of sensors and actuators for tracking and actuation; $(iii)$ integrating the sensory information in order to identify real-time actuation strategies for diverse shape-morphing tasks is complex -- the control law needs to account for the sensor \& actuation error as well as the geometric non-linearity along the entirety of the device. Traditionally, some of these issues were bypassed via extrinsic sensing~\cite{thuruthel_soft_2019,feliu-talegon_advancing_2025,tapia_makesense_2020}, using external cameras, motion trackers, etc., for pose estimation -- while effective in lab settings, they are not suitable in confined environments. Recent advances integrate thin and flexible sensors (strain gauges, optical fibers, fluidic self-sensing or capacitive skins) directly within the soft robots~\cite{zhou_integrated_2024, zou_retrofit_2024}, enabling more robust sensing -- however, they are as yet limited to sensing only the distal end of the device. A radical approach makes sensing and actuation intrinsic to the soft body~\cite{joshi_sensorless_2023,choe_soft_2023,jing_self-sensing_2024,zhang_magnetic_2025} -- mirroring biology, where muscular deformation and sensing are often spatially integrated~\cite{Boyle2012,Pehlevan2016}. The challenge still remains in making these adaptable for shape morphing tasks relevant in a variety of applications.

Inspired by these parallels, in this article, we study the effects of simple feedback strategies on the morphology of a slender soft-robot with intrinsic sensing and actuating capability. Experimental systems with embedded sensors are often limited in their sensing and actuation by the fixed number of sensors and actuators -- which is captured in our model via a filter that integrates the sensed deformation and the applied actuation. We look at the simple task of a soft device maintaining a straight posture by compensating for its self-weight by applying actuating internal moments. Based on the number of sensors and actuators, we study four regimes corresponding to combinations of sparse and high density of sensors and actuators. We show that when the applied actuation is proportional to the sensed deformation, the device undergoes a hierarchy of instabilities which result in abrupt changes in the device shape as we tune the actuation gain. The morphology of this soft system lies in the four-dimensional phase space defined by the gravito-bending number (set by the material property of the soft device), non-dimensional sensing/feedback gain (associated with sensing/feedback strategy), and the scaled filter widths (arising out of a finite number of sensors and actuators). In order to understand the limitations set by the fixed number of sensors and actuators for complex shape-morphing tasks, we compute the shape error between the achieved and target shapes, given a perfect model of the device with a limited number of sensors and actuators. We find that the error in the achieved shape is minimum for a scaled width of the filter being proportional to the applied feedback gain for a fixed gravito-bending number. Our results offer new insights into soft devices that perform diverse shape changes with limited sensing and actuating capabilities. However, our theoretical framework can be generalized to accommodate design principles for soft systems in various other applications.

\begin{figure}[t!]
\centering
\includegraphics[width=0.3\textwidth]{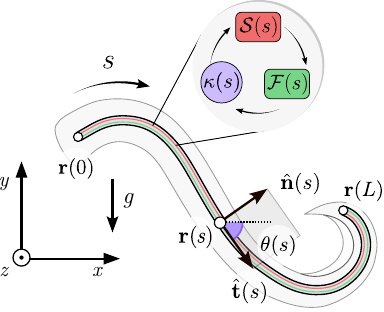}
\caption{\textbf{\textsf{Actuating elastica with sensing and feedback}} Schematic illustration of a slender soft-robot modeled as an \textit{elastica}. The centerline of the device is parameterized by the arc length $s$, with position vector $\r(s)$. The device is fixed at one end $\r(0)$ and free at the other $\r(L)$ with gravity pointing in the negative $y$-direction. Its local tangent and normal directions are denoted by $\that(s)$ and $\nhat(s)$, respectively, with the tangent vector making an angle $\theta(s)$ with the horizontal. Along the length of the device, a series of distributed sensors (red) and actuators (green) are embedded. The sensors detect the local curvature $\kappa(s)$ in the form of a sensing function $\S(s)$, which is then used to compute the feedback $\F(s)$ and the actuating moment $m_a(s)$ that drives the actuators to morph the shape of the device.} 
\label{fig:fig0}
\end{figure}

\section{\label{Model}Model}
\noindent Slender soft robots in three dimensions have three modes of deformation when they are inextensible -- two associated with bending deformation, and one with torsional deformation. The mechanics of such a slender device is described by the Kirchhoff rod equations~\cite{Audoly_2010, o2017kirchhoff, gazzola2018forward},
\begin{subequations}
\begin{align}
    \frac{\d}{\d s}\m(s) + \that(s) \times \f(s) + \q(s) = 0, \label{eq:momentBalance} \\
    \frac{\d }{\d s} \f(s) + \p(s) = 0, \label{eq:forceBalance}
\end{align}
\end{subequations}
where $\m(s)$ is the internal moment and $\f(s)$ is the internal force. Here $\q(s)$ and $\p(s)$ denote the distributed external moment and the body force per unit length along the length of the device.

Since the primary goal in this article is to understand the limitations set by the sensing and actuation modalities on the morphing capabilities of a soft robot, we confine ourselves to a planar description and model the soft robot as an \emph{elastica} of length $L$, which has only the bending mode of deformation. Our device is described in two dimensions by its centerline using the coordinates $\r(s) = (x(s), y(s))$ (see \hyperref[fig:fig0]{Fig.~\ref*{fig:fig0}}) with $s\in[0, L]$ denoting the arc length along the device. The unit tangent along the centerline of the device is $\that(s) = \d\r / \d s = (\cos \theta(s), \sin \theta(s))$ where $\theta(s)$ is the angle to the horizontal and the curvature is $\kappa(s)=\d\theta/\d s$. As the device is constrained to deform within the plane, the moment simplifies to $\mathbf{m}(s) = m(s)\ez$, where $\ez$ is the unit vector normal to the plane (shown in \hyperref[fig:fig0]{Fig.~\ref*{fig:fig0}}). 

In our description, the soft robot is able to perform shape morphing tasks using a series of sensors and actuators placed along its centerline. The sensors provide information about the local curvature via the sensing function, $\S(s) \sim f(\kappa(s))$, while the actuators can exert active local moments, $m_a(s)$. Under this setting, the scalar bending moment $m(s)$ is comprised of two distinct components: a passive elastic component $m_p(s)$ and an actuating moment $m_a(s)$, such that $m(s) = m_p(s) + m_a(s)$. The passive moment $m_p(s) = B_p\,\kappa(s)$ arises directly from the inherent stiffness of the material, where $B_p$ is the bending stiffness of the elastica. The curvature information in the sensing function $\S(s)$ is used to arrive at a feedback strategy in the form of a feedback function $\F(s)$ and the feedback translates to actuating moment $m_a(s)$, so that the soft robot can be made to transform into a particular shape (shown in \hyperref[fig:fig0]{Fig.~\ref*{fig:fig0}}). This process of sensing the current state of the device, $\kappa(s)$ via sensing function $\S(s)$ to determine the actuating moment $m_a(s) \sim g(\F(s))$ is analogous to the sensation, perception and action cycle seen in biological systems that helps them perform a variety of tasks~\cite{chelakkot_growth_2017}.

\begin{figure*}[t!]
\centering
\includegraphics[width=\textwidth]{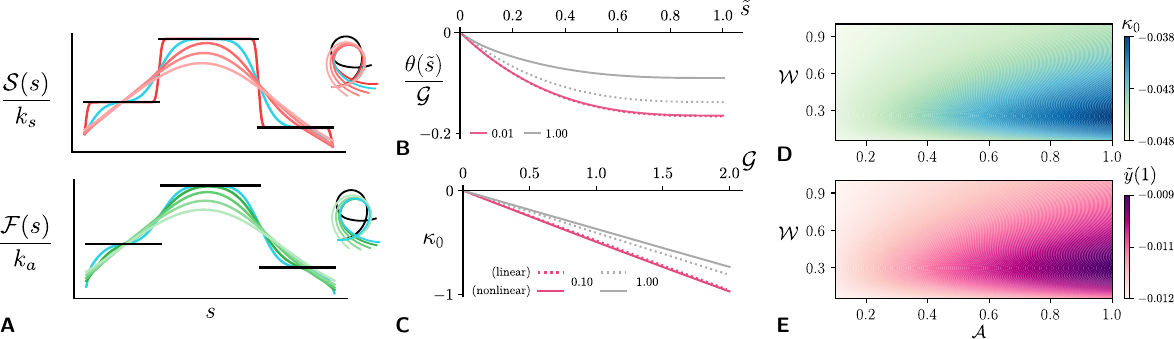}
\caption{\textbf{\textsf{Fixed number of sensors/actuators and linear solution}} (A) The effects of the number of sensors and actuators are captured in our model via Gaussian filters (see Eq.~\eqref{eq:sensing} \& \eqref{eq:feedback}) of widths $\sigma, \gamma$. Here we show the effects of tuning the width $\sigma, \gamma$ of the kernel on the sensing function, $\S(s)$, and feedback function, $\F(s)$. For a curvature profile $\kappa(s)$ of the form of two‑steps (shown in black), the normalized sensing function $\S(s)/\lambda$ (\textit{top}) obtained from Eq.~\eqref{eq:sensing} is plotted against $s$ for different widths $\sigma$. The solid red curve with the smallest $\sigma$ closely follows the target curvature, while progressively wider kernels (red with decreasing opacity) deviate increasingly. For the sensing profile highlighted in blue, we compute the feedback function, $\F(s)$ (\textit{bottom}). The normalized feedback function $\F(s)/k_a$ from Eq.~\eqref{eq:feedback} is shown for different kernel widths $\gamma$. A small $\gamma$ yields a feedback profile that best approximates the sensing input, whereas a larger $\gamma$ leads to increasing deviation. Insets depict the corresponding device shapes for each $\S(s), \F(s)$, along with the reference shape corresponding to the $\kappa(s)$ (shown in black). (B) Comparison between the linearized solution $\theta(\tilde{s})/\G$ as a function of normalized arc-length $\tilde{s}$ obtained from the series expansion in Eq.~\eqref{eq:seriesExpansion} and the nonlinear solution to Eq.~\eqref{eq:Eq_NSLA} for two non-dimensional actuation gains, $\A = 0.01$ and $\A = 0.1$ (with $\G=0.5$). For weak actuation ($\A=0.01$), the linear solution matches with the nonlinear solution, whereas at stronger actuation ($\A=0.1$) the linear solution deviates from the nonlinear. (C) Fixed-end curvature $\kappa_0 = \kappa(0)$ plotted against $\G$ for both the linearized and nonlinear formulations. As in (B), the two solutions match for small feedback strength ($\A=0.01$), while significant deviations appear at $\A=1.0$, where nonlinear effects dominate the response. (D, E) Contour map of non-dimensional $\kappa_0$ and free-end displacement $\tilde{y}(1)$ in the parameter space $(\W,\,\A)$ obtained from the linear solution. Regions of high $\kappa_0$ correspond to strong bending induced by the feedback, and increasing $\A$ also enhances the overall deflection, while larger $\W$ smoothens the response by spatially averaging the feedback. We, however, see a non-monotonic response as a function of $\W$ for a fixed $\A$, highlighting that there exists a specific width at which the device exhibits maximum response (discussed in the main text).}
\label{fig:fig1}
\end{figure*}

\subsubsection*{Sensing and feedback kernels}
\noindent We assume that each sensor module can measure local strain, which is used to infer the approximate curvature at a particular location. These discrete curvature measurements can be filtered/smoothed to obtain approximate curvature along the length of the device. The width $\sigma$ of the kernel $G_s(s,s')$ used to filter the curvature $\kappa(s)$ directly reflects the spatial resolution of the sensing arrangement: a smaller kernel width ($\sigma \ll L$), corresponds to a higher density of sensors and finer resolution, while a larger width ($\sigma \sim L$) represents a sparser distribution of sensors that leads to more spatial averaging. In the limiting case of an infinite number of densely packed sensors, the kernel $G_s(s, s')$ approaches $\delta(s-s')$, recovering sensing along the entire length of the device (here $\delta(s)$ is the Dirac delta function). In contrast, when a limited number of sensors are available, the sensing kernel becomes wide ($\sigma \rightarrow L$), enabling only an averaged measurement along the device.

Similar to the sensor configuration along the device, we also assume distributed actuators along the spatial extent. The actuators generate moments $m_a(s)$ that are spread over finite segments of length $\gamma$. To mathematically capture this effect, we model actuation as a nonlocal process defined over a spatial extent using a kernel $G_a(s, s')$. The width $\gamma$ of the actuation kernel directly reflects the spatial extent of the actuators: a small kernel width ($\gamma \ll L$) corresponds to highly localized actuation capable of applying targeted moments over a small region, whereas a large kernel width ($\gamma \sim L$) represents broadly distributed actuation that produces large-scale deformations. In the limiting case $\gamma \rightarrow 0$, the actuation becomes local, recovering point-wise application of moment. Conversely, for $\gamma \rightarrow L$, the actuation tends toward a uniform global input applied across the entire structure.

In this framework, the actuation moment, $m_a(s)$, is derived directly from the sensed curvature $\S(s)$, making the feedback effectively proportional to $\kappa(s)$. This feedback rule is similar to proportional control, where the control input depends on the error~\cite{astrom_feedback_2021}. Similar to biological systems, which rely on proprioceptive sensing to detect their shape~\cite{chelakkot_growth_2017}, the sensing function captures the curvature information by integrating over a spatial region via the kernel $G_s(s, s^\prime)$. The sensing function, $\S(s)$ is given by
\begin{align}
    \S(s) = \lambda \int_0^L G_s(s,s^\prime)\, \kappa(s^\prime) \, \d s^\prime,
    \label{eq:sensing}
\end{align}
where, $\lambda$ is the sensing gain and the kernel $G_s(s, s^\prime)$ characterizes the spatial resolution of the sensing mechanism. We assume that the kernel has translational invariance and takes the form $G_s(s-s^\prime) = \exp\!\left[-(s-s^\prime)^2/(2\sigma^2)\right]/\sqrt{2\pi\sigma^2}$, where $\sigma$ is the width of the Gaussian kernel, which determines the size of the sensing region. Smaller $(\sigma/L)$ corresponds to a narrow kernel and highly localized sensing, and in the limit $\sigma \to 0$, the sensing function reduces to $\S(s) = \lambda \kappa(s)$, which retrieves the state of the device via local sensing. In contrast, when $\sigma$ is large, the kernel spans a significant portion of the device, leading to spatial averaging of curvature. In the extreme case where $(\sigma/L) \rightarrow 1$, the sensing becomes global and simplifies to $\S(s) = \lambda[\theta(L)-\theta(0)]$, independent of arc length. \hyperref[fig:fig1]{Fig.~\ref*{fig:fig1}A} (Top) shows the curvature profile (black lines), the spatial variation of normalized sensing function $\S(s)/\lambda$, for different values of $\sigma$, along with the corresponding shape of the device. We see that for $(\sigma/L) \ll 1$, the device's curvature profile gets reconstructed accurately. 

Since we have access only to the sensing function $\S(s)$, our feedback function $\F(s)$ for actuation must depend on $\S(s)$ alone. Taking into account the finite size effects of the actuator described earlier, the feedback function $\F(s)$ is given by,
\begin{align}
    \F(s) = \beta \int_0^L G_a(s,s^\prime) \, \mathcal{S}(s^\prime) \, \d s^\prime,
    \label{eq:feedback}
\end{align}
where $\beta$ is the feedback gain and $G_a(s,s^\prime)$ is the actuation kernel. We again assume that the non-local actuation kernel, $G_a(s-s^\prime) = \exp\!\left[-(s-s^\prime)^2/(2\gamma^2)\right]/\sqrt{2\pi\gamma^2}$ with width $\gamma$, capturing the finite-size effect.

As with sensing, the parameter $\gamma$ controls the spatial distribution of actuators: a narrow kernel, $(\gamma/L) \ll 1$, corresponds to dense actuators capable of localized control, whereas a wide kernel, $(\gamma/L) \rightarrow 1$, arises from sparse actuation, resulting in spatially averaged moment generation. \hyperref[fig:fig1]{Fig.~\ref*{fig:fig1}A} (Bottom) shows the spatial variation of normalized function $\F(s)/\beta$ for different $\gamma$ values, along with the corresponding device shapes. This feedback function $\F(s)$ can be directly used to determine the actuating moment, $m_a(s)$ and in our model, we assume that $m_a(s)\equiv\F(s)$. In the local limit $\gamma \to 0$, the feedback reduces to $m_a(s) = \lambda\beta \, \S(s)$, where the actuation at each location depends only on the local sensing signal $\S(s)$. On the other hand, for $(\gamma/L) \rightarrow 1$, the actuation becomes spatially uniform and depends only on the net shape change of the device, $\F = \beta[\S(L)-\S(0)]$.

\subsubsection*{Straightening a sagging Elastica}

\noindent Assuming the device experiences no distributed external moment, $\q(s)=0$, and that no external force acts at its free end, $\f(L)=0$, the force balance Eq.~\eqref{eq:forceBalance} yields the internal force components $f_x(s) = 0$ and $f_y(s) = -\rho g (L - s)$, where $\rho$ is the linear mass density and $g$ is the acceleration due to gravity. Substituting these into the moment balance Eq.~\eqref{eq:momentBalance}, we arrive at the governing equation,
\begin{align}
    B_p \frac{\d^2\theta}{\d s^2} + \frac{\d m_a}{\d s} - \rho g (L - s) \cos\theta = 0.
    \label{eq:Eq_motion}
\end{align}
Together with the geometric relations, $\d x/\d s = \cos\theta$ and $\d y/\d s = \sin\theta$, Eq.~\eqref{eq:Eq_motion} defines the equilibrium shape of the device as a function of the local angle $\theta(s)$ and the actuating moment $m_a(s)$ generated through feedback. In the special case where the actuating moment vanishes ($m_a=0$), Eq.~\eqref{eq:Eq_motion} reduces to the classical elastica equation describing a passive inextensible beam under its self-weight in a gravitational field~\cite{Audoly_2010}. In what follows, we explore the task of the sagging device becoming straight by accounting for its self-weight. In order to perform this task, we look at simple strategies wherein the sensor information, $\S(s)$, is used to compute the actuating moment, $m_a(s)$. We explore the effects of the sensing and the feedback gains, $k_s, k_a$, on the morphology of the device with prescribed sensing and actuating capabilities, set by the kernel widths $\sigma, \gamma$.

\begin{figure*}[t!]
\centering
\includegraphics[width=\textwidth]{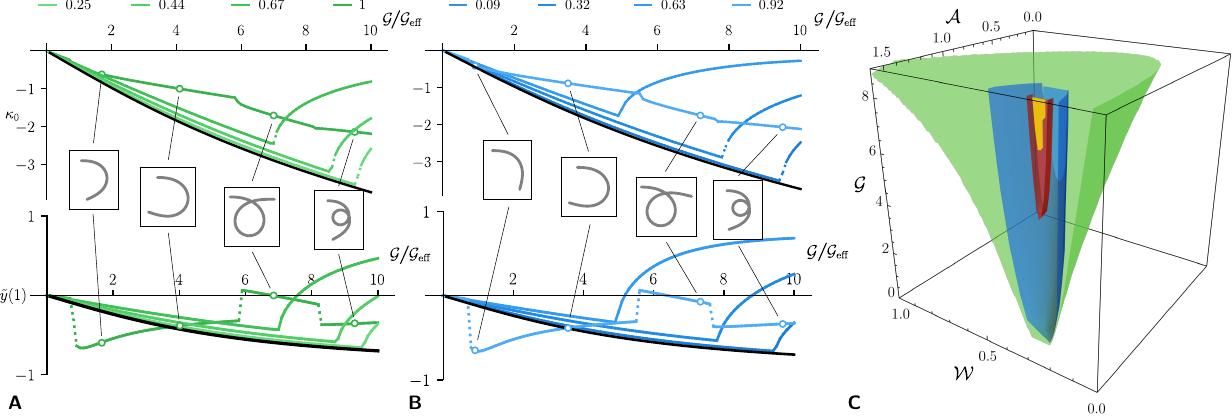}
\caption{\textbf{\textsf{Sensing/Feedback gain driven morphological instability}} (A) Bifurcation diagrams ($\kappa_0, \tilde{y}(1)$ vs $\G$) for fixed actuation strength $\A = 1.0$ and different sensing width $\W = 0.25,\,0.44,\,0.67,\,1.00$. Discontinuities in the selected branch for $\W=0.25$ highlight shape transitions, with representative device morphologies displayed as insets. (B) Bifurcation diagrams for fixed sensing width $\W = 0.23$ and different non-dimensional actuation gain $\A = 0.09,\,0.32,\,0.63,\,0.92$. The inset shows the equilibrium shapes for the strongest actuation ($\A = 0.92$). In both (A) and (B), the black curve denotes the local limit $\kappa_0, \tilde{y}(1)$ vs $\G_\eff$ for comparison (discussion in main text). (C) Three-dimensional phase-diagram captured by the gravito-bending number $\G$, the scaled sensing width $\W$, and the non-dimensional actuation gain $\A$ scaled by $\W$. We see four bifurcation surfaces that delineate distinct morphologies of the soft device. We see changes in the equilibrium shapes across these surfaces, corresponding to transitions in $\kappa_0$ and $\tilde{y}(1)$.}
\label{fig:fig2}
\end{figure*}

\section{\label{Results}Results}

\noindent To systematically investigate the influence of sensing function, $\S(s)$ and actuation function, $\F(s)$ on the effective execution of the task by the device, we now consider four distinct scenarios characterized by different combinations of the normalized widths $\sigma/L, \gamma/L$ of the sensing and actuation kernels,  $G_s(s, s')$ and $G_a(s, s')$. To solve the nonlinear system described by Eq.~\eqref{eq:Eq_motion} accurately (along with the geometric relations), we employ the \textit{Chebyshev spectral collocation method}~\cite{Trefethen_2000, guo_application_2012} and the non-local integrals in these equations are evaluated using \textit{Clenshaw-Curtis} integration scheme~\cite{clenshaw_method_1960,boyd_chebyshev_2001}. The spectral technique offers high accuracy and exponential convergence with the number of grid-points for smooth solutions, thus capturing the nonlinear deformations of the device efficiently (see SI Sec.~\ref{sec:methodSI} for details). The four scenarios we study in this article are $(i)$ \textit{local sensing and actuation}, the asymptotic case with a very large number of densely packed sensors and actuators that result in both sensing and actuation kernel widths being sharply localized i.e., $(\sigma, \gamma) \rightarrow 0$; $(ii)$ \textit{global sensing and actuation}, wherein the sensors and actuators are available only at the extremities of the device, i.e. $(\sigma, \gamma) \sim L$; $(iii)$ \textit{mixed sensing and actuation}, involving small number of sensors(/actuators) combined with densely packed actuators(/sensors), i.e. $\sigma$ or $\gamma \sim L$; $(iv)$ \textit{nonlocal sensing and actuation}, which is the general case with a small number of sensors and actuators corresponding to both sensing and actuation kernels of nonzero widths, i.e. $0 < (\sigma, \gamma) < L$.

\subsection{Local sensing and actuation, $(\sigma, \gamma) \rightarrow 0$}
\label{subsec:LSLA}
\noindent In the limit when $(\sigma, \gamma) \rightarrow 0$ in $G_s(s, s')$ and $G_a(s, s')$, the resulting actuating moment, $m_a(s)$ is directly proportional to the local curvature, $m_a(s) = \lambda \beta \kappa(s)$. Substituting this into Eq.~\eqref{eq:Eq_motion} and non-dimensionalizing the equation by the device length $L$ (see SI Sec.~\ref{subsec:LSLA_SI} for details), we obtain the governing equation,
\begin{align}
    \frac{\d^2\theta(\tilde{s})}{\d \tilde{s}^2} - \G_\eff \,(1-\tilde{s})\,\cos\theta(\tilde{s}) = 0, 
    \label{eq:Eq_LSLA}
\end{align}
along with the geometric relations $\d\tilde{x}/\d\tilde{s} = \cos\theta(\tilde{s})$ and $\d\tilde{y}/\d\tilde{s} = \sin\theta(\tilde{s})$, where $\tilde{s}=s/L$, $\tilde{x}=x/L$, and $\tilde{y}=y/L$. As the device is clamped at the base ($\tilde{s}=0$) and free at the distal end ($\tilde{s}=1$), it satisfies the boundary conditions: $\theta(0)=0, \d\theta(1)/\d\tilde{s}=0, \tilde{x}(0)=\tilde{y}(0)=0$.  The behavior of the system in this local sensing and actuation limit is governed by a \emph{single dimensionless parameter}, $\G_\eff = \rho g L^3/(B_p + \lambda\beta)$. Physically, $\G_\eff$ quantifies the relative strength of gravity to the \emph{effective bending stiffness} $(B_p + \lambda\beta)$, which includes both the passive material stiffness $B_p$ and the additional stiffening from sensing and feedback gains $\lambda\beta$. 

For small values of~$\G_\eff$, the device remains nearly horizontal with minimal bending. As $\G_\eff$ increases, curvature near the clamped end increases, leading to pronounced downward deflection. In the $\G_\eff \gg 1$ regime, nonlinear effects dominate, resulting in sharp curvature gradients concentrated near the base corresponding to the significant sagging. This transition from small-deflection to large-deflection behavior is captured by two key observables: the curvature at the clamped end, $\kappa_0 = \kappa(0)$, and the vertical deflection of the free end, $\tilde{y}(1)$. The variation of $\kappa_0$ and $\tilde{y}(1)$ as a function of $\G_\eff$ is shown by black curves in {\hyperref[fig:fig1]{Fig.~\ref*{fig:fig2}A}} and {\hyperref[fig:fig1]{\ref*{fig:fig2}B}} (see also SI video~\ref{SI_video1} for variation of $\kappa_0$ and $\tilde{y}(1)$ as a function of $\G_\eff$). We see that by increasing the actuation strength, $k_s k_a$ (or equivalently decreasing~$\G_\eff$), the device flattens toward a straight configuration, and for larger $k_s k_a$, the device curls upwards, away from the sagged state. This limit serves as a baseline for understanding how a finite number of sensors and actuators alter the capabilities of morphing the shape of the slender device.

\subsection{Global sensing and actuation, $(\sigma, \gamma) \rightarrow L$}
\label{subsec:NSNA_uniform}
\noindent When $(\sigma, \gamma) \to L$, the sensed curvature and actuation depend only on the \emph{global shape change} rather than local curvature variations. In this regime, the actuating moment reduces to a spatially uniform value, $m_a = (\beta \lambda/L^2)\big[\theta(L)-\theta(0)\big]$, capturing the difference between the distal and proximal angles. The boundary conditions in this regime change to $\d\theta(1)/\d \tilde{s} = \alpha$, where $\alpha$ is the prescribed actuating moment at the free end. It is worth noting that in the global sensing and actuation regime, we have two independent non-dimensional parameters: applied distal moment, $\alpha$ and bendo-gravity number, $\G=\rho g L^3 / B_p$.

Since this regime maps exactly to the problem of the shape of a hair discussed in detail in ref.~\cite{Audoly_2010}, we limit to a brief description of the results (ref. SI Sec.~\ref {subsec:NSNA_uniform_SI} for more details). \hyperref[fig:SI_fig_1]{Fig.~\ref*{fig:SI_fig_1}} summarizes the dependence of the clamped-end curvature $\kappa_0$ and free-end deflection $\tilde{y}(1)$ on these parameters. For a fixed applied moment $\alpha$, increasing $\G$ (i.e., stronger gravity or weaker bending stiffness) reduces the curvature at the clamped end $\kappa_0$ and increases the downward deflection of the free end (see \hyperref[fig:SI_fig_1]{Fig.~\ref*{fig:SI_fig_1}A}). However, unlike the local case, the free-end curvature is not set to zero (because of the applied moment, $\alpha$ at $\tilde{s}=1$); even for large $\G$, the prescribed $\alpha$ induces a curling at the tip. Stronger actuation (corresponding to larger $\alpha$) globally stiffens the structure, mitigating sagging. For a fixed $\G$, $\kappa_0$ increases monotonically with $\alpha$, as the global actuation amplifies the base curvature (see \hyperref[fig:SI_fig_1]{Fig.~\ref*{fig:SI_fig_1}B}). In contrast, the free-end deflection $\tilde{y}(1)$ exhibits a non-monotonic response: $(i)$ For small $\alpha$, increasing the global gain initially reduces sagging (lifting the tip); $(ii)$ Beyond a critical $\alpha$, the uniform feedback introduces curling at the free end, causing the tip deflection to reverse; $(iii)$ For very large $\alpha$, the system tends toward a globally straightened shape.

The non-monotonic dependence of free-end deflection $\tilde{y}(1)$ on tip actuation strength $\alpha$ reveals a key physical distinction between local and uniform global feedback. Local feedback behaves like a distributed stiffening proportional to the beam’s local curvature. In contrast, global feedback redistributes actuation moments across the entire structure, enabling long-range modulation of internal forces. This mechanism can initially amplify bending near the free end before eventually suppressing sag for sufficiently strong actuation.

\subsection{Mixed sensing and actuation, $0 < \sigma\;\text{or}\;\gamma < L$}
\label{subsec:NSLA_mixed}
\noindent In the intermediate regime between the local and global sensing and actuation limits, we have one of the sensing or the actuation kernels remaining local, i.e., $\sigma$ or $\gamma \to 0$, and the other width being nonzero. This leads to the actuating moment, $m_a(s)$ of the form,
\begin{align*}
    m_a(s) = \beta \lambda \int_0^L \frac{1}{\sqrt{2\pi\sigma^2}} \exp\!\left(-\frac{(s - s^\prime)^2}{2\sigma^2}\right) \frac{\d \theta}{\d s^\prime} \, \d s^\prime.
\end{align*}
We see that the feedback at any point depends on a weighted average of curvature along the device, with $\sigma$ defining the width of the filter. Substituting this actuating moment into Eq.~\eqref{eq:Eq_motion} and introducing the non-dimensional arc-length $\tilde{s}=s/L$, the equilibrium shape is governed by the nonlinear integro-differential equation (see SI Sec.~\ref{subsec:NSLA_SI} for details)
\begin{align}
    \begin{split}
        \frac{\d^2 \theta}{\d\tilde{s}^2} 
        + \frac{\A}{\sqrt{2\pi}\,\W} \frac{\d}{\d\tilde{s}}
            \left[
                \int_0^L \exp\!\left(-\frac{(\tilde{s} - \tilde{s}^\prime)^2}{2\W^2}\right) \frac{\d\theta}{\d\tilde{s}^\prime} \, \d\tilde{s}^\prime
            \right] \\
        \;-\; \G (1 - \tilde{s}) \cos \theta = 0. 
        \label{eq:Eq_NSLA}
    \end{split}
\end{align}
This equation involves three independent non-dimensional parameters that together dictate the device's behavior: $(i)$ Bendo-gravity number, $\G = \rho g L^3 \,/\, B_p$ which we have already seen in Sec.~\ref{subsec:LSLA}; $(ii)$ Non-dimensional gain, $\A = \lambda\beta \,/\, B_p$ which measures the relative strength of the internal sensing–actuation feedback compared to passive stiffness. $(iii)$ Scaled kernel width, $\W=(\sigma/L)$ is the characteristic sensing length scale that defines the spatial extent over which curvature is averaged. $\A$ and $\W$ together control the magnitude and spatial range of the feedback, while $\G$ represents the external driving load. It should be pointed out that interchanging the kernel roles -- setting $\sigma \to 0$ (local sensing) and $\gamma$ finite (nonlocal actuation) -- leads to exactly the same governing equation in Eq.~\eqref{eq:Eq_NSLA}. This implies that the device has the same utility with a fixed number of sensors or actuators.\\

\begin{figure*}[t!]
\centering
\includegraphics[width=0.95\textwidth]{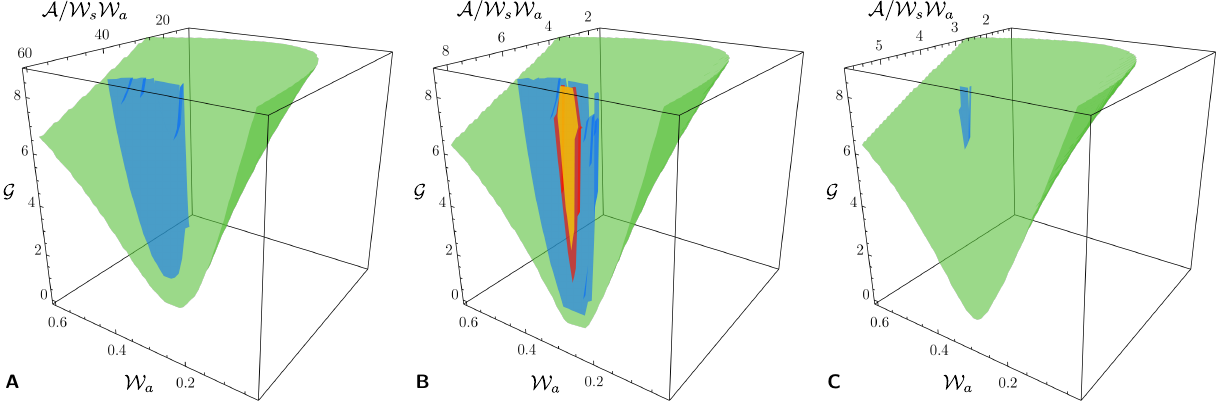}
\caption{\textbf{\textsf{Phase-space with non-local sensing and actuation.}} The nonlocal limit of sensing and actuation has four non-dimensional numbers, bendo-gravity number $\G$, the scaled sensing width $\W_s=(\sigma/L)$, the scaled actuation width $\W_a=(\gamma/L)$ and the effective actuation gain $\A/(\W_s\W_a)$. (A) For $\W_s=0.02$, bifurcations are sparse and appear only at very large actuation gains ($\A \approx 55$), indicating that extremely localized actuation requires disproportionately high gain to trigger instabilities. (B) For $\W_s=0.2$, several bifurcations are observed within a moderate range of $\A$ (up to $\A \approx 9$), suggesting enhanced sensitivity when actuation is distributed over an intermediate length scale. (C) For $\W_s=0.5$, only a few bifurcations are observed because the kernel acts as a strong spatial low-pass filter, destabilizing only the long wavelength modes. In all cases, the phase diagrams are computed for $\W_a \in [0.01,0.6]$ and $\G \in [0.1,8]$.}
\label{fig:fig3}
\end{figure*}

\noindent \textit{Limit of $\G, \A \ll 1$:}
In the limit of small bendo-gravity number and non-dimensional gain, $\G, \A \ll 1$, the effects of gravity as well as the activating moment are small in Eq.~\eqref{eq:Eq_NSLA}. In this regime, we can linearize this equation while retaining the nonlocal term. The series solution to this equation takes the form,
\begin{align}
    \theta(\tilde{s}) = \sum_{n=0}^\infty c_n \sin\!\left(\frac{(2n+1)\pi \tilde{s}}{2}\right),
    \label{eq:seriesExpansion}
\end{align}  
where the chosen basis function satisfies the boundary conditions $\theta(0)=0$ and $\d\theta(1) \,/\, \d \tilde{s} = 0$ (see SI Sec.~\ref{subsec:NSLA_series_SI} for further details about the series solution). After solving for $c_n$ in terms of $\G, A$ using the linearized equation, we can evaluate the base curvature $\kappa_0$ and the transverse tip displacement $\tilde{y}(1)$ as a function of non-dimensional gain, $\A$ and the kernel width, $\W$ (shown in \hyperref[fig:fig1]{Fig.~\ref*{fig:fig1}D, E}). The feedback gain $\A$ uniformly amplifies the response: increasing $\A$ causes larger magnitudes of the base curvature $\kappa_0$ and larger tip deflections $\tilde{y}(1)$ before eventual saturation. On the other hand, the kernel width $\W$ has a non-monotonic effect on the device response, as the maximum response is seen for intermediate $\W$ values. Physically, this observation can be attributed to the fact that a narrow kernel acts point-wise and fails to capture curvature variations over the characteristic deformation length scale of the device determined by $\G$. In contrast, a very wide kernel width $\W$ distributes the signal over nearly the entire body, which effectively averages out spatial details and thereby attenuates the actuation input. At the intermediate values of $\W$, the kernel captures curvature variations over a relevant length scale, resulting in more effective feedback, and the observables $\kappa_0$ and $\tilde{y}(1)$ attain the largest values. In the nonlinear problem discussed below, the same mechanism promotes a richer bifurcation phenomenon observed for intermediate values of $\W$.

\subsubsection*{Actuation driven morphological transitions}  
\noindent In the nonlinear regime characterized by $\G, \A \geq 1$, we find that the device undergoes a sequence of morphological transitions as we tune the non-dimensional actuation gain, $\A$, and the kernel width, $\W$. \\

\noindent\textit{Varying $\W$ at fixed $\A$ \& varying $\A$ at fixed $\W$:} \hyperref[fig:fig2]{Fig.~\ref*{fig:fig2}A} shows the fixed-end curvature $\kappa_0$ and the free-end displacement $\tilde{y}(1)$, as a function of $\G$ for different values of $\W$. For small sensing widths ($\W=0.25$), both $\kappa_0$ and $\tilde{y}(1)$ exhibit multiple discontinuous transitions, and at each transition, the device undergoes a change in its morphology (shown in the inset of \hyperref[fig:fig2]{Fig.~\ref*{fig:fig2}A}). As we tune $\G$, the device progressively acquires a complex shape, and when $\W \rightarrow 1$ (tending towards global sensing), these transitions become continuous. In \hyperref[fig:fig2]{Fig.~\ref*{fig:fig2}B} we present similar plots for four different values of $\A$ at fixed $\W$. For weak actuation ($\A=0.09$), the response of the system is continuous with shape change (shown in the inset), and as $\A$ increases, multiple discontinuous transitions emerge ($\A=0.92$). In both \hyperref[fig:fig2]{Figs.~\ref*{fig:fig2}A, B}, the black curves show $\kappa_0, \tilde{y}(1)$ vs $\G_\eff$ from the local limit (corresponding to $(\sigma, \gamma) \to 0$), serving as a reference. These curves show that the behavior of the device in the mixed sensing and actuation is similar to the local case with $\G_\eff$ when $\A/\W < 1$. We can see that the Eq.~\eqref{eq:Eq_NSLA} resembles Eq.~\eqref{eq:Eq_LSLA} but with $\G$ instead of $\G_{\eff}$ when the effect of $\A/\W$ is small. For deviations in $\A/\W$ from this limit, we see changes in the device behavior from the local scenario. Further, discontinuous transitions appear when the effects of the nonlocal term in Eq.~\eqref{eq:Eq_NSLA} start playing an important role. SI~Videos~\ref{SI_video2} and~\ref{SI_video3} illustrate these bifurcations, the first by varying the sensing width $\W$ at fixed $\A=1.6$ and $\G=8$, and the second by varying the actuation strength $\A$ at fixed $\G=8$ and $\W=0.22$.

These transitions are encapsulated in the form of a three-dimensional phase diagram parameterized by $(\G,\,\W,\,\A/\W)$ (shown in \hyperref[fig:fig2]{Fig.~\ref*{fig:fig2}C}) which provides the overall morphological space of the device. The ratio $\A/\W$, which quantifies the strength of the nonlocal feedback relative to the sensing width, is the appropriate axis for visualization (see SI.~\ref{subsec:NSLA_SI} for details), and phase boundaries represented by four surfaces partition the parameter space into distinct device morphologies (shown in the inset of \hyperref[fig:fig2]{Figs.~\ref*{fig:fig2}A, B}).

\begin{figure*}[t!]
\centering
\includegraphics[width=0.95\textwidth]{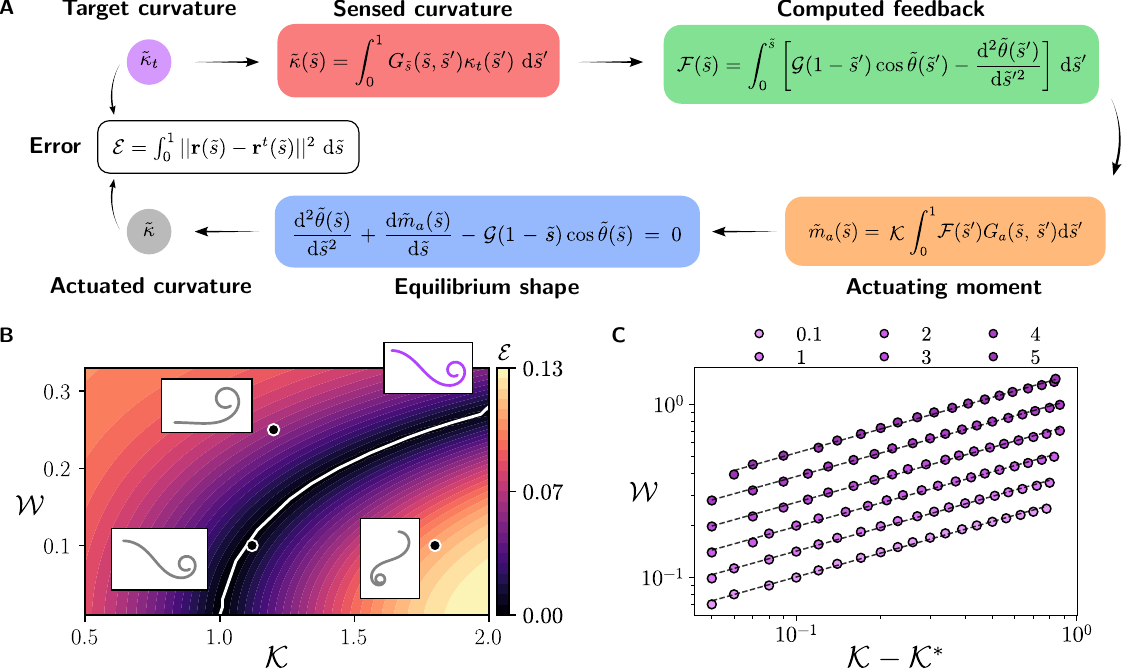}
\caption{\textbf{\textsf{Shape morphing with limited sensing \& feedback}} (A) Algorithm for shape morphing used in simulation, where for a given target curvature, $\kappa^t(\tilde{s})$, we compute the sensed curvature $\tilde{\kappa}(s)$, which is a filtered version of the target curvature with kernel $G_s(s, s^\prime)$. Using the sensed curvature and a perfect model of the device, we can evaluate the feedback required for actuation, $\mathcal{F}(s)$. This feedback is filtered to result in the actuating moment $\tilde{m}_a(s)$, owing to the finite number of actuators by a kernel $G_a(s, s^\prime)$ with $\K$ being a non-dimensional actuation gain. The equilibrium shape of the device for this applied actuating moment is $\tilde{\kappa}(\tilde{s})$ with shape error $\mathcal{E}$. (B) Contour plot of the shape error $\mathcal{E}$ as a function of the actuation gain $\K$ and kernel width $\W$. A distinct minimum-error region (dark blue) appears, indicating optimal $(\K,\,\W)$ pairs that best reproduce $\kappa^t(\tilde{s})=3(\tilde{s}+1)^3-(2\tilde{s}-3)^2$). Four representative shapes in the inset show close agreement within the minimum-error region and clear deviations away from it. (C) Log–log plot of optimal scaled sensing width, $\W^*$ as a function of $(\K - \K^*)$ for several values of the gravito-bending number, $\G$ (individual curves manually shifted vertically for clarity) reveals $\W^* \sim (\K - \K^*)^{\zeta}$ with $\zeta \approx 0.5$.}
\label{fig:fig4}
\end{figure*}

\subsection{Non-local sensing and actuation, $0 < (\sigma, \gamma) < L$}
\label{subsec:NSNA_General}
\noindent We now explore the most general feedback configuration, where the width of both $G_a(s, s')$ and $G_s(s, s')$ lies in the intermediate regime, i.e., $0 < (\sigma, \gamma) < L$, representing a small number of sensors and actuators. The four non-dimensional parameters governing this behavior of the device are (see SI sec.~\ref{subsec:NSNA_SI} for details): $(i)$~$\G = \rho g L^3\,/\, B_p$, $(ii)$~$\A = \beta\lambda\,/\, B_p$, $(iii)$~$\W_s=(\sigma/L)$, the scaled sensing kernel width, $(iv)$~$\W_a=(\gamma/L)$, the scaled actuation kernel width.\\

\noindent \textit{Bifurcations and morph-space:} The equilibrium morphology of the device in this regime retains key features of the mixed sensing and actuation seen earlier. We see multiple bifurcations characterized by shape transitions, and transition towards complex shapes as we explore the morphological space of the device in the phase-space parameterized by $(\G,\,\A,\,\W_s,\,\W_a)$. The solution space becomes significantly richer and also computationally intensive to capture. To make the exploration tractable, we fix the actuation width $\W_s$ at three representative values -- $\W_s=0.02$, $\W_s=0.1$, and $\W_s=0.5$ -- and construct \textit{3D phase diagrams} in $(\G,\,\W_a,\,\A/\W_s\W_a)$ for each $\W_s$. The resulting phase-diagrams are shown in \hyperref[fig:fig3]{Fig.~\ref*{fig:fig3}}.

When sensing is almost local ($\W_s=0.02$), we observe that the device gradually straightens with an increase in $\A/\W_s\W_a$ and then bends upward smoothly, with one bifurcation appearing at extremely large actuation gain (shown in \hyperref[fig:fig3]{Fig.~\ref*{fig:fig3}A}). In the intermediate sensing width ($\W_s=0.1$), multiple bifurcations appear, closely resembling the rich morphological space observed in the mixed sensing and actuation case (ref. \hyperref[fig:fig3]{Fig.~\ref*{fig:fig3}B}). Last, in the large sensing width ($\W_s=0.5$), the device again undergoes shape change under smooth transition as $\A/\W_s\W_a$ increases (as seen in \hyperref[fig:fig3]{Fig.~\ref*{fig:fig3}C}). At sufficiently high $\A/\W_s\W_a$ values, two isolated bifurcations still appear, leading to upward-bending and coiling of the device with higher modes of deformation.

Thus, while the general case retains the fundamental bifurcation features observed in the mixed sensing and actuation case, the inclusion of a second spatial kernel enriches the parameter space. From a practical perspective, we see that a small number of sensors and actuators in soft robots need to be carefully designed to either suppress undesired instabilities or deliberately exploit multi-stability for reconfigurable shape control.
\subsection{Shape morphing into complex shapes}
\label{subsec:limit_actuation}
\noindent So far, we have looked at the simple task of a slender device actuating to compensate for its self-weight to remain straight. We see that even in such a simple task, the device can undergo actuation-driven shape transitions. In practice, soft robotic systems often have to perform complex shape morphing tasks while equipped only with a small number of sensors and actuators. A key question that arises from this practical constraint is: how well can a device morph its shape given a fixed number of sensors and actuators? Further, can we identify an appropriate gain for the actuators such that the device can transform to complex target shapes with minimum error? 

To address these questions, we introduce a framework for shape morphing (illustrated schematically in \hyperref[fig:fig4]{Fig.~\ref*{fig:fig4}A}) where the slender device has to morph to a target shape described by curvature, $\kappa_t(\tilde{s})$. Accounting for the finite number of sensors, the target curvature is sensed as,
\[
\tilde{\kappa}(\tilde{s}) = \int_0^1 G_s(\tilde{s}, \tilde{s}') \tilde{\kappa_t}(\tilde{s}') \ \d \tilde{s}'.
\]
Using this sensed curvature, we can compute the ideal feedback moment required to achieve this sensed curvature via the equilibrium condition in Eq.~\eqref{eq:Eq_motion} to get
\[
\F(\tilde{s}) = \int_0^{\tilde{s}} \bigg[ \G(1-\tilde{s}')\cos \theta(\tilde{s}') - \frac{\d^2 \theta(\tilde{s}')}{\d \tilde{s}'^2} \bigg] \ \d \tilde{s}'.
\]
It is important to note that we assume here that the moment applied along the device can be tuned based on the feedback $\F(\tilde{s})$ (unlike in the previous cases where this was set proportional to $\S(\tilde{s})$), however with a spatial resolution set by kernel width, $\W = \sigma/L$). To reflect this constraint set by the finite size of actuators, the calculated feedback is filtered again through an actuation kernel to give,
\[
    \tilde{m}_a(\tilde{s}) = \K \int_0^1 G_a(\tilde{s},\tilde{s}^\prime) \, \F(\tilde{s}^\prime) \, \d \tilde{s}^\prime.
\]

The shape error between the new equilibrium state of the device $\r(\tilde{s})$ under this filtered feedback and the target shape $\r^t(\tilde{s})$ can be written as
\[
\E[\W, \K; \G]
~:=~
\int_0^{1} 
\big\| \mathbf{r}(\tilde{s}) \;-\; \mathbf{r}^t(\tilde{s}) \big\|^2
\,\mathrm{d}\tilde{s}.
\]
We now put this framework to test for a target shape with a long-wavelength and a short-wavelength feature by choosing $\kappa_t(\tilde{s}) = 3\,(\tilde{s}+1)^3 - (2\tilde{s}-3)^2$ (shown in the inset of \hyperref[fig:fig4]{Fig.~\ref*{fig:fig4}B} in purple). By systematically varying the actuation gain $\K$ and the kernel width $\W$, we evaluate the error landscape $\mathcal{E}(\K,\,\W)$ (shown in \hyperref[fig:fig4]{Fig.~\ref*{fig:fig4}B}). Large $\E$  corresponds to either capturing the long-wavelength feature or the short-wavelength feature inaccurately. We find a minimum-error band emerges in the $(\K,\,\W)$--plane (see SI Video~\ref{SI_video4} for an illustration) corresponding to optimal combinations of actuation strength and sensor \& actuator width $\W$ that best reproduce the target shape. In contrast, moving away from the minimum-error region corresponds to either capturing the long-wavelength feature or the short-wavelength feature inaccurately (shown as insets in \hyperref[fig:fig4]{Fig.~\ref*{fig:fig4}B}), leading to deviations away from the target shape.

We look at the locus of the kernel width, $\W^*(\K)$ corresponding to minimal error,
\[
\W^*(\K) \,:=\, \textrm{argmin}_{\W} \, \E[\W, \,\K].
\]
When $\W=0$, this corresponds to local sensing and feedback wherein the configuration of the device, $\tilde{\kappa}(\tilde{s})$ can be sensed by $\S(\tilde{s})$ perfectly and the computed feedback $\F(\tilde{s})$ must result in perfect matching of the target shape (as seen in \hyperref[fig:fig4]{Fig.~\ref*{fig:fig4}B}). However, when $W \neq 0$, the filter width influences the capability of the shape morphing. We find that the optimal width, $\W^*(\K)$ has a simple power-law dependence on $\K$, shown in \hyperref[fig:fig4]{Fig.~\ref*{fig:fig4}C}. We find that $\W^* \sim (\K - \K^*)^{\zeta}$ on logarithmic axes for different values of the gravito-bending number, $\G$. Here $\K^*$ is the critical gain required to overcome gravity, and from a linear fit between $\W^*, \K$ we find $\zeta \approx 0.5$. In order to minimize the shape error $\E[\W,\, \K]$ with nonzero $\W$, the gain $\K$ can be tuned so that there is an appropriate trade-off between capturing the long-wavelength as well as the short-wavelength features with a perfect model.

By selecting a $(\K,\,\W^*)$ pair along the determined minimum shape error line, one can determine the optimal gain of the sensors and actuators that can make the device morph to complex shapes. This insight can directly inform the design of distributed sensing-actuation with simple feedback strategies in soft robots, where trade-offs between hardware complexity and control accuracy are crucial.

\section{Conclusion}
\label{sec:conclusion}
In this article, we have investigated a continuum model of a slender soft-robot with spatially distributed curvature sensing and moment-generating actuation. By exploring the effects of limited sensing and actuating capabilities, we have identified a morphological space of the device in the form of a phase diagram. The diagram has a physical attribute along one axis (via the gravito-bending number), actuating gain along another axis, and the spacing between sensors/size of actuators along the third. We see that the device can undergo morphological transitions as we traverse this phase space, capturing the sensitivity of the device to the physical tuning parameters. Moreover, for complex shape morphing tasks, we find an important trade-off between accurate shape reproduction and system stability with a simple proportional feedback. Wider sensing distributions improve stability by smoothing local variations, but at the cost of reduced accuracy in reproducing fine features. In contrast, higher actuation gains enhance responsiveness but can lead to overcompensation, introducing unintended curvature or instabilities. Our work highlights the limitations set by simple feedback strategies in a non-linear system with limited sensing and actuation, and also provides a quantitative framework to make informed design choices under such constraints.\\

\noindent \textbf{Acknowledgments}: We thank IIT Madras and ANRF-ECRG/2024/003341/ENS for partial financial support. We also thank the INTERFACE lab members for their interactions. The data and the SI Videos are available at \href{https://github.com/sgangaprasath/ActuatingElastica2025/}{Github}.
\bibliographystyle{apsrev4-1}
\bibliography{Elastica}

\widetext
\clearpage
\onecolumngrid
\begin{center}
\textbf{\large Supplemental Information for \\[.2cm]
``Morphologies of a sagging \textit{elastica} with intrinsic sensing and actuation''}\\[.2cm]
Vishnu  Deo  Mishra$^{1}$, S  Ganga  Prasath$^{1}$\\[.1cm]
{\small \itshape ${}^1$Department of Applied Mechanics \& Biomedical Engineering, IIT Madras, Chennai, TN 600036.\\
}
\end{center}
\setcounter{equation}{0}
\setcounter{figure}{0}
\setcounter{table}{0}
\setcounter{page}{1}
\setcounter{section}{0}
\makeatletter
\setstretch{1.5}

\renewcommand{\thetable}{S\arabic{table}}%
\renewcommand{\thesection}{S\arabic{section}}
\renewcommand{\thesubsection}{SS\arabic{subsection}}
\renewcommand{\theequation}{S\arabic{equation}}
\renewcommand{\thefigure}{S\arabic{figure}}

\section{SI videos}
\noindent All supplementary videos can be accessed directly by clicking on their respective links below.
\subsection{Video~1: Local sensing and actuation, $(\sigma, \gamma) \rightarrow 0$}
\label{SI_video1}
\noindent \href{https://drive.google.com/file/d/1CjOyJteWA9-tqVybe_Gmtz3j2rkg6Cbh/view?usp=sharing}{Video 1}: This video shows three linked visualizations for the local feedback case:  
(1) the curvature at the clamped end $\kappa_0$ versus $\G$,  
(2) the vertical deflection of the free end $\tilde{y}(1)$ versus $\G$,  
and (3) the instantaneous shape of the elastica as $\G$ is varied interactively.  
Both $\kappa_0$ and $\tilde{y}(1)$ vary linearly for small $\G$ and gradually saturate for large $\G$ due to geometric non-linearities. At the same time, the shape visualization highlights the transition from strongly sagging to nearly straight configurations.

\subsection{Video~2: Morphological transitions with $\W$ in mixed sensing and actuation, $\sigma$ or $\gamma \rightarrow 0$}
\label{SI_video2}
\noindent \href{https://drive.google.com/file/d/19Q51BEEaBzraYFWjoyCCirO-Lv0t3m2R/view?usp=sharing}{Video 2}: This video illustrates the evolution of the device morphology as the normalized sensing width $\W$ is varied for fixed actuation strength $\A=1.6$ and gravito-bending number $\G=8$. The left panel shows the fixed-end curvature $\kappa_0$ and free-end displacement $\tilde{y}(1)$ plotted against $\W$. In both plots, transitions in the response are highlighted with dotted lines, indicating bifurcation points where the equilibrium shape switches between distinct solution branches.  

The right panel displays the instantaneous equilibrium shape of the device corresponding to the current value of $\sigma$. At the bifurcation points, the device undergoes a sudden morphological transition.

\subsection{Video~3: Morphological transitions with $\A$ in mixed sensing and actuation, $\sigma$ or $\gamma \rightarrow 0$}
\label{SI_video3}
\noindent \href{https://drive.google.com/file/d/1swSvqZBQqHZ9_A5NafFQd88q8aaqIv3q/view?usp=sharing}{Video3}:
This video demonstrates how the equilibrium morphology of the device evolves as the actuation strength $\A$ is varied while keeping the gravito-bending number fixed at $\G=8$ and the normalized sensing width at $\W=0.22$. The left panel tracks the variation of fixed-end curvature $\kappa_0$ and free-end displacement $\tilde{y}(1)$ as functions of $\A$. Transition in these curves, marked by dotted segments, indicates bifurcation points corresponding to qualitative changes in equilibrium states. The right panel simultaneously shows the instantaneous device shapes associated with the current value of $\A$. With increasing $\A$, the device goes through a sequence of distinct configurations, with switching at the bifurcation points.

\subsection{Video~4: Shape morphing by tuning $\K, \,\W$}
\label{SI_video4}
\noindent \href{https://drive.google.com/file/d/1om42raI9ZheD_5QfkKlTfBCUQg55KPEU/view?usp=sharing}{Video 4}: This video dynamically illustrates the error landscape and its effect on shape reproduction. The left panel shows the contour plot of the error $\mathcal{E}$ between the computed shape and the target one in the $(\K,\,\W)$--plane. A marker traverses continuously along a predefined path across the contour plot. At each marker position, the corresponding parameter values $(\K,\,\W)$ are used in the two-stage sensing-actuation framework (described in the main text) to compute the final equilibrium shape of the device.  

The right panel simultaneously displays the computed device shape (gray curve) and the fixed target shape (orange curve). As the marker approaches the minimum-error region in the contour plot, the device aligns closely with the target, showing minimal deviation. Moving away from this optimal band gradually increases the mismatch. This clearly demonstrates how the interplay between actuation gain $\K$ and kernel width $\W$ directly impacts the precision of shape tracking, emphasizing the existence of an optimal sensing-actuation footprint.  

\section{Limits of sensing and actuation}
\subsection{Local sensing \& Local Actuation, $(\sigma, \gamma) \rightarrow 0$}\label{subsec:LSLA_SI}
\noindent In this section, we analyze the limiting case of local sensing and local actuation, which corresponds to the asymptotic regime $\sigma \to 0$ and $\gamma \to 0$, and derive the governing equation of motion. Here, $\sigma$ and $\gamma$ denote the spatial widths of the sensing and actuation kernels in Eqs.~\eqref{eq:sensing} and \eqref{eq:feedback}, respectively. Physically, this limit represents an idealized \emph{point-wise feedback} scenario, in which both the sensing and actuation are restricted to infinitesimally small neighborhoods along the device. In this regime, the Gaussian kernel collapses to Dirac delta functions, i.e.,
\[
G_s(s,s^\prime) \;\to\; \delta(s - s^\prime) 
\quad \text{and} \quad
G_a(s,s^\prime) \;\to\; \delta(s - s^\prime),
\]
which implies that the feedback depends only on the \emph{local curvature} at the same arc-length location. Substituting these kernels into the sensing function~\eqref{eq:sensing} yields
\begin{align*}
    \S(s) &= \lambda \int_0^L \delta(s - s^\prime) \kappa(s^\prime) \, \d s^\prime = \lambda \, \kappa(s),
\end{align*}
where $\kappa(s) = \d\theta / \d s$ is the curvature. Substituting this into the actuation function~\eqref{eq:feedback} gives 
\begin{align*}
    \F(s) &= \beta \int_0^L \delta(s - s^\prime) \S(s^\prime) \, ds^\prime = \beta \, \lambda \, \kappa(s).
\end{align*}
Thus, in this regime, the \emph{active bending moment} is simply proportional to the local curvature, $m_a(s) = \F(s) = \beta \lambda \kappa(s)$.

Substituting this feedback-induced moment into the moment balance equation~\eqref{eq:Eq_motion} gives
\begin{align*}
    \big(B_p + \beta \lambda\big) \, \frac{\d^2 \theta}{\d s^2} - \rho g (L - s) \cos \theta &= 0.
\end{align*}

To express this equation in a non-dimensional form, we rescale the arc length as $\tilde{s} = s/L$ and define the dimensionless geometric coordinates $\tilde{x} = x/L$ and $\tilde{y} = y/L$. We introduce a non-dimensional parameter, the effective gravito-bending number, defined as $\G_\eff = \rho g L^3/(B_p + \beta\lambda)$. This parameter measures the strength of gravitational loading relative to the \emph{effective stiffness}, incorporating both passive elasticity $B_p$ and active feedback $\beta\lambda$ of the system. With this rescaling, the governing equation becomes 
\begin{align}
    \frac{\d^2 \theta(\tilde{s})}{\d\tilde{s}^2}  - \G (1-\tilde{s}) \cos\,\theta(\tilde{s}) = 0.\label{SI_eq:Eq_LSLA}.
\end{align}
The associated geometric relations remain $\d \tilde{x} / \d\tilde{s} = \cos\theta(\tilde{s})$, and $\d \tilde{y} / \d\tilde{s} = \sin\theta(\tilde{s})$,  along with the augmented boundary conditions, $\theta(0)=0,\,\d\theta(1)/\d \tilde{s}=0,\,\tilde{x}(0)=0=\tilde{y}(0).$

\subsection{Nonlocal sensing and actuation, $(\sigma, \gamma) \rightarrow L$}
\label{subsec:NSNA_uniform_SI}
\noindent In this asymptotic scenario, both the sensing and actuation kernels are nonlocal, spanning the entire device. Mathematically, this corresponds to $\sigma, \gamma \rightarrow L$ in Eqs.~\eqref{eq:sensing} and~\eqref{eq:feedback}. In this limit, the Gaussian kernels lose their spatial localization and reduce to a uniform distribution normalized over the domain:
\begin{align*}
    G_s(s,s^\prime), \, G_a(s,s^\prime) \; \longrightarrow \; \frac{1}{L}.
\end{align*}
With this uniform kernel, the sensed curvature signal becomes the \emph{average curvature} along the device:  
\begin{align*} 
    \S(s) &= \lambda \int_0^L G(s, s^\prime) \kappa(s^\prime) \, \d s^\prime = \frac{\lambda}{L} \left[ \theta(L) - \theta(0) \right]. 
\end{align*}
Hence, in the nonlocal limit, all points along the device receive the same sensing signal proportional to the net angular change between the two ends. Since $\S(s)$ is spatially uniform, the corresponding actuation term simplifies to
\begin{align*} 
    \F(s) &= \beta \int_0^L K(s, s^\prime) \S(s^\prime) , \d s^\prime = \frac{\beta\lambda}{L^2} \left[ \theta(L) - \theta(0) \right]. 
\end{align*}
Thus, the actuating moment $m_a(s) \equiv \mathcal{F}(s)$ is spatially constant, determined solely by the \emph{total angular displacement} of the device. Substituting this expression into the equation of motion Eq.~\eqref{eq:Eq_motion}, we obtain
\begin{align*}
    B_p \frac{\d^2 \theta}{\d s^2} + \frac{\d}{\d s}\left(\frac{\beta\lambda}{L^2} \big[ \theta(L) - \theta(0) \big] \right) - \rho g (L - s) \cos \theta &= 0.
\end{align*}
Since $\big[\theta(L)-\theta(0)\big]$ is a constant, differentiation eliminates the actuation term, reducing the above equation to
\begin{align*}
    B_p \frac{\d^2 \theta}{\d s^2} - \rho g (L - s) \cos \theta &= 0.
\end{align*} 
Introducing the dimensionless arc length $\tilde{s}=s/L$ and dividing through by $B_p/L^2$, we obtain:
\begin{align} 
    \frac{\d^2 \theta(\tilde{s})}{\d \tilde{s}^2} - \G \, (1-\tilde{s})\cos\theta(\tilde{s}) = 0, 
    \label{SI_eq:NSNA_uniform}
\end{align}
where the dimensionless parameter is the gravito-bending number, $\G = \rho g L^3/B_p$ -- captures the relative magnitude of gravitational loading with respect to the passive bending stiffness.

Unlike the local case, the uniform nonlocal feedback influences the system through an additional boundary constraint. In addition to the boundary condition $\theta(0) = 0$, since the actuation depends explicitly on the end angles, the free-end slope must satisfy a prescribed value, $\d\theta (1) /\d\tilde{s} = \alpha$. Here, $\alpha$ is a free parameter controlling the degree of active curling at the free end. Physically, $\alpha$ quantifies how the global feedback modulates the device’s boundary rotation through long-range coupling between all segments.
\begin{figure*}[t!]
\centering
\includegraphics[width=0.95\textwidth]{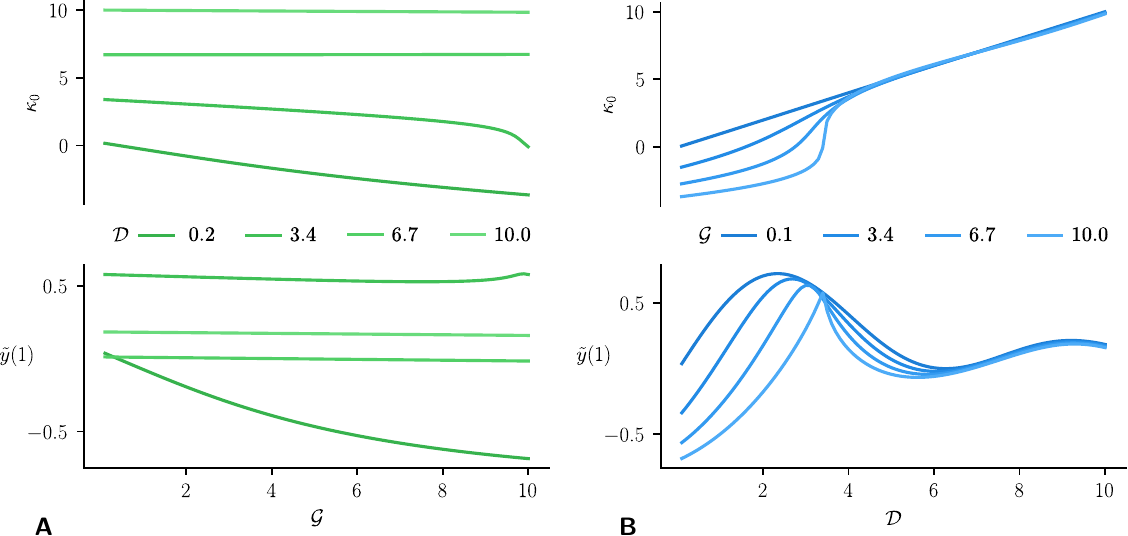}
\caption{\textbf{\textsf{Shape morphing under uniform global feedback.}} 
(A) Bifurcation diagram for different applied end moment $\alpha = 0.1,\,3.4,\,6.7,\,10.0$ as a function of gravito-bending number $\G$. The top and bottom panels show the variation of clamped-end curvature $\kappa_0$ and free-end deflection $\tilde{y}(1)$, respectively. As $\G$ increases (stronger gravity or weaker stiffness), the $\kappa_0$ decreases and the free end sags more deeply. Unlike the local-feedback case, the prescribed $\alpha$ prevents the tip curvature from vanishing, so a residual ``curling'' remains even at large $\G$. Increasing $\alpha$ globally stiffens the beam, mitigating sag and elevating both $\kappa_0$ and $\tilde{y}(1)$. (B) Bifurcation diagram for different gravitational load $\G = 0.5,\,2.5,\,5.0,\,7.5$ as a function of $\alpha$. The top panel shows the variation of clamped-end curvature $\kappa_0$, which increases monotonically with $\alpha$ for all the $\G$ values. The free-end deflection $\tilde{y}(1)$ (bottom panel), on the other hand, shows a non-monotonic trend: increasing $\alpha$ at first lifts the tip, reducing sagging of the device. However, beyond a critical $\alpha$, the uniform actuation begins to overcompensate, resulting in additional tip curling that reverses the direction of deflection, before eventually straightening out again for sufficiently large $\alpha$ (see section~\ref{subsec:NSNA_uniform} for a detailed discussion).}
\label{fig:SI_fig_1}
\end{figure*}

Thus, in the uniform nonlocal limit, the bulk shape equation~\eqref{SI_eq:NSNA_uniform} remains the same as the classical device under gravity, but the boundary conditions are augmented by $\alpha$, introducing an additional degree of freedom. Consequently: (\romannumeral 1) $\G$ controls the overall sagging under gravity (similar to the local case), and (\romannumeral 2) $\alpha$ introduces active free-end bending, enabling equilibrium shapes with controlled free-end curvature even when the distributed feedback is uniform.

\subsection{Mixed sensing and feedback, $0 < \sigma\;\text{or}\;\gamma < L$}
\label{subsec:NSLA_SI}
\noindent Here, we derive the governing equation for the mixed case where sensing is nonlocal while actuation remains strictly local. This scenario is particularly relevant to practical systems, where sensors typically collect curvature information over a finite spatial region, but actuators apply feedback locally. Mathematically, this regime corresponds to a finite sensing width $\sigma \neq 0$ in Eq.~\eqref{eq:sensing}, while taking $\gamma \to 0$ in Eq.~\eqref{eq:feedback}. Thus, the sensing function is represented by a Gaussian kernel $G_s(s,s^\prime)$ that integrates curvature over a neighborhood of characteristic length $\sigma$. The actuation, however, is still applied pointwise, captured by a Dirac delta kernel $G_a(s,s^\prime) = \delta(s-s^\prime)$.  

The sensed curvature in this scenario at a point $s$ is given by
\begin{align*}
\S(s) &= \lambda \int_0^L G(s, s^\prime) \kappa(s^\prime) \, \d s^\prime = \lambda \int_0^L \frac{1}{\sqrt{2\pi\sigma^2}} \exp\!\left(-\frac{(s - s^\prime)^2}{2\sigma^2}\right) \frac{\d\theta}{\d s^\prime} \, \d s^\prime,
\end{align*}
where the Gaussian kernel $G(s,s^\prime)$ averages the sensed curvature over a finite sensing length $\sigma$. Since the actuation is local, the feedback torque applied at $s$ is  
\begin{align*}
\F(s) &= \beta \int_0^L \delta(s - s^\prime) \S(s^\prime) , \d s^\prime = \beta \lambda \int_0^L \frac{1}{\sqrt{2\pi\sigma^2}} \exp\!\left(-\frac{(s - s^\prime)^2}{2\sigma^2}\right) \frac{\d\theta}{\d s^\prime} \, \d s^\prime.
\end{align*} 
Thus, the actuating moment $m_a(s)$ is a Gaussian-weighted average of local curvature, modulated by both the sensing gain $\lambda$ and actuation gain $\beta$.  

Substituting $m_a(s)=\F(s)$ into the general moment balance Eq.~\eqref{eq:Eq_motion}, we obtain  
\begin{align*}
B_p \frac{\d^2 \theta}{\d s^2} + \frac{\d}{\d s}\left[\beta \lambda \int_0^L \frac{1}{\sqrt{2\pi\sigma^2}} \exp\!\left(-\frac{(s - s^\prime)^2}{2\sigma^2}\right) \frac{\d\theta}{\d s^\prime} \, \d s^\prime\right] - \rho g (L - s) \cos \theta &= 0.
\end{align*}
Introducing non-dimensional variables $\tilde{s} = s\,/\,L$, $\W = \sigma\,/\,L$, and dividing through by the bending stiffness $B_p$, the governing equation becomes 
\begin{align}
    \frac{\d^2 \theta}{\d\tilde{s}^2} + 
    \frac{\A}{\sqrt{2\pi}\,\W} \frac{\d}{\d\tilde{s}}\left[\int_0^L \exp\!\left(-\frac{(\tilde{s} - \tilde{s}^\prime)^2}{2\W^2}\right) \frac{\d\theta}{\d\tilde{s}^\prime} \, \d\tilde{s}^\prime\right] - 
    \G (1 - \tilde{s}) \cos \theta &= 0,
    \label{SI_eq:NSLA} 
\end{align}
where,  
\begin{align*}
    \A = \frac{\beta\lambda}{B_p},\;\;\; \text{and} \;\;\;\G = \frac{\rho g L^3}{B_p}. 
\end{align*}
The parameter $\A$ quantifies the strength of curvature-feedback relative to passive stiffness, and gravito-bending number $\G$ measures the gravitational loading relative to bending stiffness, controlling the tendency of the device to sag under its own weight. The width $\W$ determines the spatial distribution of the sensing kernel, interpolating between local ($\W\to 0$) and global sensing ($\W\to 1$).  

\noindent \textit{On the choice of $\A/\W$ in phase diagrams~\hyperref[fig:fig2]{Fig.~\ref*{fig:fig2}C}.} The feedback contribution in Eq.~\ref{SI_eq:NSLA} enters as a Gaussian filtering followed by a spatial derivative, which introduces a prefactor proportional to $1/\W$. Thus, the effective feedback strength naturally scales with the ratio $\A/\W$ and reflects the actuation gain in the governing equation. Therefore, the three-dimensional phase diagram providing the overall device stability is created using $\A,\,\A/\W,\,\text{and }\W$ as the parameter set.

\noindent \textit{Indistinguishable role of $\sigma$ and $\gamma$ in nonlocality.} An important observation in the mixed case is that if we swap the roles of sensing and actuation -- taking $G_s(s,s^\prime)=\delta(s-s^\prime)$ (local sensing) while $G_a(s,s^\prime)$ is a Gaussian kernel with finite width $\gamma\neq 0$ -- the resulting actuating moment
\[
m_a(s) = \beta \int_0^L 
\frac{1}{\sqrt{2\pi\gamma^2}}\,
\exp\!\!\left(-\frac{(s-s^\prime)^2}{2\gamma^2}\right)\,
\kappa(s^\prime)\,\d s^\prime
\]
leads to exactly the same governing equation \eqref{SI_eq:NSLA} after non-dimensionalization (with $\gamma$ playing the role of $\sigma$). This mathematical symmetry implies that the system cannot distinguish whether the nonlocality stems from sensing or actuation -- only the integrated spatial extent of feedback matters. From an experimental perspective, this equivalence can be exploited in designing the soft robotic limbs. 

\subsection{Nonlocal sensing and nonlocal feedback, $0 < (\sigma,\,\gamma)<L$}
\label{subsec:NSNA_SI}
\noindent In this case, we move to a scenario where both the sensing and actuation processes are non-local, which corresponds to the nonzero $\sigma$ and $\gamma$ values so that Gaussian kernels model the corresponding operations. Specifically, the sensing function is given by a Gaussian kernel $G_s(s,s^\prime)$ that smooths the curvature over a spatial scale $\sigma$, while the actuation function is modeled by another Gaussian kernel $G_a(s,s^\prime)$ with a characteristic length $\gamma$. These kernels account for the finite distributed nature of both the sensing and actuation along the device. 

The sensed curvature in such a scenario is expressed as
\begin{align*}
\S(s) &= \lambda \int_0^L G_s(s, s^\prime) \,\kappa(s^\prime) \, \d s^\prime = \lambda \int_0^L \frac{1}{\sqrt{2\pi\sigma^2}}\, \exp\!\left(-\frac{(s - s^\prime)^2}{2\sigma^2}\right) \frac{\d\theta}{\d s^\prime} \, \d s^\prime,
\end{align*}
where the kernel $G(s,s^\prime)$ effectively averages the local curvature $\kappa(s^\prime)= \d\theta / \d s^\prime$ over a region determined by $\sigma$. Since the feedback is also nonlocal, the feedback moment is obtained by convolving the sensed curvature with the Gaussian kernel $K(s,s^\prime)$ and is expressed as
\begin{align*}
\F(s) &= \beta\int_0^L G_a(s, s^\prime) \, \S(s^\prime) \, \d s^\prime \\
      &= \beta \lambda \int_0^L \left[ \frac{1}{\sqrt{2\pi\gamma^2}} \exp\!\left(-\frac{(s - s^\prime)^2}{2\gamma^2}\right) \int_0^L \left\{ \frac{1}{\sqrt{2\pi\sigma^2}} \exp\!\left(-\frac{(s^\prime - s^{\prime\prime})^2}{2\sigma^2}\right) \frac{\d\theta}{\d s^{\prime\prime}} \right\} \d s^{\prime\prime} \right] \d s^\prime.
\end{align*} 

Upon non-dimensionalizing with $\tilde{s}=s/L$, $\W_s=\sigma/L$, and $\W_a=\gamma/L$, the governing equation becomes 
\begin{align}
    \frac{\d^2 \theta}{\d\tilde{s}^2} + \frac{\A}{2\pi \W_s \W_a} \frac{\d}{\d\tilde{s}}
    \left[
        \int_0^L \biggl\{ \exp\!\left(-\frac{(\tilde{s} - \tilde{s}^\prime)^2}{2\W_a^2}\right) \int_0^L \exp\!\left(-\frac{(\tilde{s}^\prime - \tilde{s}^{\prime\prime})^2}{2\W_s^2}\right) \frac{\d\theta}{\d\tilde{s}^{\prime\prime}} \d\tilde{s}^{\prime\prime}\biggl\}\, \d\tilde{s}^\prime
    \right] 
    - \G (1 - \tilde{s}) \cos \theta &= 0, \label{SI_eq:NSNA_finite}
\end{align}
where,
\begin{align*}
    \A = \frac{\beta\lambda}{B_p},\;\;\; \text{and} \;\;\;\G = \frac{\rho g L^3}{B_p}.
\end{align*}
The dimensionless parameter $\A$, in this case, encapsulates the coupling between the distributed sensing (through $\lambda$ and $\sigma$) and the nonlocal actuation (through $\beta$ and $\gamma$). This reflects the fact that, in many practical systems, sensors and actuators may not operate over identical spatial extents. The parameter $\G$ measures the relative influence of gravitational loading on the system, normalized by the bending stiffness.

\section{Small-Angle Approximation: Local Case}
\label{SI:small_angle}
\noindent Under the small-angle assumption, $\cos\,\theta(\tilde{s}) \approx 1$, the nonlinear equation governing the local limit Eq.~\eqref{eq:Eq_LSLA} in the main text reduces to the linear ordinary differential equation,
\begin{align}
     \frac{\d^2 \theta}{\d\tilde{s}^2} - \G (1 - \tilde{s}) = 0. \label{eq:linear_local}
\end{align}
Integrating once with respect to $\tilde{s}$ and imposing the boundary condition $\d\theta\,/\,\d\tilde{s}\,(\tilde{s}=1) = 0$ and $\theta(0) = 0$, leads to
\begin{align*}
    \theta(\tilde{s}) = \G\left(\frac{\tilde{s}^2}{2} - \frac{\tilde{s}^3}{6} - \frac{\tilde{s}}{2}\right).
\end{align*}

\section{Series Solution: Local Case}
\label{SI:series_solution}
\noindent For further validation and analytical insight in the local case, we also construct a series solution to the small-angle linearized equation~\ref{eq:linear_local}. We seek a solution satisfying clamped-free boundary conditions $\theta(0)=0$ and $\d\theta\,/\,\d\tilde{s}\,(\tilde{s}=1)=0$. To ensure these conditions are automatically satisfied, we expand the solution on a sine basis:
\begin{align}
    \theta(\tilde{s}) = \sum_{n=0}^{\infty} a_n \sin\left(\frac{(2n+1) \pi \tilde{s}}{2}\right). \label{eq:series_expn_SI}
\end{align}
The corresponding second derivative is:
\begin{align}
    \frac{\d^2 \theta}{\d\tilde{s}^2} = -\sum_{n=0}^{\infty} a_n \left(\frac{(2n+1)\pi}{2}\right)^2 \sin\left(\frac{(2n+1) \pi \tilde{s}}{2}\right). 
    \label{eq:dd_series_SI}
\end{align}
To express the forcing term $\G(1 - \tilde{s})$ in the same sine basis, we write:
\begin{align}
    \G(1 - \tilde{s}) = \sum_{n=0}^{\infty} b_n \sin\left(\frac{(2n+1) \pi \tilde{s}}{2}\right),
    \label{eq:gravity_series_SI}
\end{align}
with the coefficients given by:
\begin{align*}
    b_n = 2 \int_0^1 \G(1 - \tilde{s}) \sin\left(\frac{(2n+1) \pi \tilde{s}}{2}\right) \d\tilde{s}.
\end{align*}
Using integration by parts, we obtain the closed-form expression:
\begin{align*}
    b_n = \frac{4\G}{(2n+1)\pi} - \frac{8\G}{((2n+1)\pi)^2} (-1)^n.
\end{align*}
Substituting the above expansions for $\d^2 \theta\,/\,\d\tilde{s}^2$ and $\G(1 - \tilde{s})$ into Eq.~\eqref{eq:linear_local} and equating coefficients results:
\begin{align}
    a_n = -\frac{16 \G}{((2n+1)\pi)^3} + \frac{32 \G}{((2n+1)\pi)^4} (-1)^n. \nonumber
\end{align}
Hence, the complete solution is:
\begin{align}
    \theta(\tilde{s}) = \sum_{n=0}^{\infty} \left[ -\frac{16 \G}{((2n+1)\pi)^3} + \frac{32 \G}{((2n+1)\pi)^4} (-1)^n \right] \sin\left(\frac{(2n+1)\pi \tilde{s}}{2}\right). \label{eq:series_final}
\end{align}

This series representation not only provides analytical insights into the gravitation-induced deformation in the small-angle, local regime, but also provides a valuable benchmark for validating the spectral numerical method discussed in Section~\ref{subsec:NSLA_mixed}, especially in the small $\G$ regime where non-linearities remain weak.

\section{Series Solution: Nonlocal Sensing \& Local Actuation}
\label{subsec:NSLA_series_SI}
\noindent We next move to the series solution of the case where sensing is nonlocal (modeled using a Gaussian kernel of width $\W$), while actuation remains strictly local. The governing equation for this scenario, in non-dimensional form, is given by Eq.~\eqref{eq:Eq_NSLA}. Under the small-angle approximation, $\cos \theta \approx 1$, the equation becomes:
\begin{align}
    \frac{\d^2 \theta}{\d\tilde{s}^2} + \frac{\A}{\sqrt{2\pi}\,\W} \frac{\d}{\d\tilde{s}}
    \left[
        \int_0^1 \exp\!\left(-\frac{(\tilde{s} - \tilde{s}^\prime)^2}{2\W^2}\right) \frac{\d\theta}{\d\tilde{s}^\prime} \, \d\tilde{s}^\prime
    \right] 
    - \G (1 - \tilde{s}) = 0. \label{eq:NSLA_SeriesSol_SI}
\end{align}

The series expansions for the first and third terms of the above equation are already derived in Eq.~\eqref{eq:dd_series_SI} and \eqref{eq:gravity_series_SI}, respectively. The remaining second term -- the convolution involving the Gaussian kernel -- can also be expanded in the same sine basis:
\begin{align*}
    \frac{\A}{\sqrt{2\pi}\,\W} \frac{\d}{\d\tilde{s}}
    \left[
        \int_0^1 \exp\!\left(-\frac{(\tilde{s} - \tilde{s}^\prime)^2}{2\W^2}\right) \frac{\d\theta}{\d\tilde{s}^\prime} \, \d\tilde{s}^\prime
    \right] 
    = \sum_{n=0}^{\infty} c_n \sin\left(\frac{(2n+1)\pi \tilde{s}}{2}\right), 
    \label{eq:NSLA_convolution_series}
\end{align*}
where the coefficients $c_n$ are given by:
\begin{align*}
    c_n = 2 \frac{\A}{\sqrt{2\pi}\,\W} 
    \int_0^1 
    \left[ 
        \int_0^1 -\left(\frac{\tilde{s} - \tilde{s}^\prime}{\W^2}\right) \exp\!\left(-\frac{(\tilde{s} - \tilde{s}^\prime)^2}{2\W^2}\right) \frac{\d\theta}{\d\tilde{s}^\prime} \, \d\tilde{s}^\prime 
    \right]
    \sin\left(\frac{(2n+1)\pi \tilde{s}}{2}\right) \d\tilde{s}. 
\end{align*}
Here, the expansion of Eq.~\eqref{eq:NSLA_SeriesSol_SI}'s 2nd term only is shown; 1st and 3rd (containing $\G$) is given by Eq.~\eqref{eq:dd_series_SI} and \eqref{eq:gravity_series_SI}. Using the series representation for the first derivative,
\begin{align*}
    \frac{\d\theta}{\d\tilde{s}^\prime} = \sum_{m=0}^{\infty} a_m \frac{(2m+1) \pi}{2} \cos\left(\frac{(2m+1) \pi \tilde{s}^\prime}{2}\right), 
\end{align*}
we can express the coefficient $c_n$ as:
\begin{align}
    \begin{split}
        c_n = \frac{\A}{\sqrt{2\pi}\,\W} \sum_{m=0}^{\infty} a_m (2m+1) \pi
        \int_0^1 
        \Bigg[ 
            \int_0^1 - \left(\frac{\tilde{s} - \tilde{s}^\prime}{\W^2}\right) \exp\!\left(-\frac{(\tilde{s} - \tilde{s}^\prime)^2}{2\W^2} \right) \cos\left(\frac{(2m+1) \pi \tilde{s}^\prime}{2}\right) \, \d\tilde{s}^\prime 
        \Bigg] 
        \\
        \sin\left(\frac{(2n+1) \pi \tilde{s}}{2}\right) \d\tilde{s}. 
    \end{split}
\end{align}
The full series solution to Eq.~\eqref{eq:NSLA_SeriesSol_SI} is then constructed by equating the sine coefficients of all three terms, which yields a linear algebraic system for the unknown coefficients $\{a_n\}$. This system can be truncated and solved numerically to approximate the device shape. Although an exact closed-form expression for $a_n$ can be derived, the resulting formula is prohibitively long and hence we do not reproduce it here. Instead, we provide the corresponding \textsc{Mathematica} notebooks used for this derivation and solution in a \href{https://github.com/sgangaprasath/ActuatingElastica2025/}{GitHub} repository. This semi-analytical approach serves as a valuable tool for understanding the system's behavior under weak non-linearity, and for benchmarking the numerical spectral scheme employed in this article.

\section{Methods}
\label{sec:methodSI}
\noindent To numerically solve the boundary value problems governing the morphology of the device under various combinations of spatially distributed sensing and actuation (Eq.~\eqref{eq:Eq_LSLA},\eqref{eq:Eq_NSLA}~\&\eqref{SI_eq:NSNA_finite}), we employ a Chebyshev spectral collocation method~\cite{Trefethen_2000,boyd_chebyshev_2001}. This technique provides high accuracy on finite domains and is well-suited to smooth solutions defined on finite intervals. In addition, we utilize Clenshaw-Curtis quadrature to efficiently approximate the nonlocal integral terms. We work on the nondimensionalized domain $\tilde{s}\in[0,1]$. The solution $\theta(\tilde{s})$ is approximated as a truncated Chebyshev expansion,
\begin{align}
    \theta(\tilde{s}) \;\approx\; \sum_{n=0}^{N} a_n \, T_n(x(\tilde{s})),
\end{align}
where $T_n(x)$ are Chebyshev polynomials of the first kind, and $x(\tilde{s}) = 2\tilde{s}-1$ maps $\tilde{s}\in[0,1]$ to $x\in[-1,1]$. The collocation points are chosen as the Chebyshev-Gauss-Lobatto nodes,
\begin{align*}
    x_j = \cos\!\left(\frac{\pi j}{N}\right), \qquad j = 0,\dots,N,
\end{align*}
which cluster near the boundaries, thereby improving the resolution of boundary-layer features. An affine transformation maps these nodes to the physical domain $\tilde{s}_j = (1+x_j)/2$. First and second-order derivatives at collocation points are approximated using standard Chebyshev differentiation matrices $D$ and $D^2$, rescaled for the interval $[0,1]$. Boundary conditions (e.g., $\theta(0)=0$, $\d\theta(1)/\d\tilde{s}=0$) are imposed by modifying or eliminating rows of the differentiation matrices so that the constraints are satisfied exactly.

The integro-differential terms in Eq.~\eqref{eq:Eq_NSLA}, which involve convolution with Gaussian kernels, are evaluated using Clenshaw-Curtis quadrature~\cite{Trefethen_2000}. Specifically, for a generic kernel $K(\tilde{s},\tilde{s}')$, the integral is approximated as
\begin{align}
    I(\tilde{s}_j) = \int_0^1 K(\tilde{s}_j,\tilde{s}') \, f(\tilde{s}') \, d\tilde{s}'
    \;\approx\; \sum_{k=0}^N w_k \, K(\tilde{s}_j,\tilde{s}_k) \, f(\tilde{s}_k),
\end{align}
where $\{w_k\}$ are Clenshaw–Curtis quadrature weights associated with the collocation nodes. This approach leverages the same Chebyshev grid for both differentiation and integration, ensuring consistency and spectral accuracy.

The resulting discretized equation is solved iteratively using the standard Newton-Raphson scheme. We initialize with $\theta(\tilde{s})=0$, and for small $\G$ (e.g., in Eq.~\eqref{eq:Eq_LSLA}), the solution is weakly nonlinear and readily computed. Solutions for larger values of $\G$ are constructed by continuation, using the converged solution at a lower $\G$ as the initial guess. 

For more complex cases involving nonlocal sensing and actuation kernels (e.g., in Eq.~\eqref{eq:Eq_NSLA}), we adopt a two-stage continuation strategy: first, the local problem is solved to convergence; next, the nonlocal parameters $(\A,\W_s,\W_a)$ are introduced gradually, with the solution at each step serving as the initial guess for the subsequent one. This incremental approach stabilizes the iterations and ensures convergence even in strongly nonlinear regimes.

\end{document}